\documentclass[conference]{IEEEtran}
\IEEEoverridecommandlockouts
\usepackage{cite}
\usepackage{amsmath,amssymb,amsfonts}
\usepackage{algorithmic}
\usepackage{graphicx}
\usepackage{textcomp}
\usepackage{xcolor}
\usepackage{booktabs}  
\usepackage{float}
\usepackage{algorithm}
\usepackage{amsmath}
\usepackage{graphicx}
\usepackage{subcaption}
\usepackage{multirow}
\usepackage{soul}
\usepackage{xcolor}
\def\BibTeX{{\rm B\kern-.05em{\sc i\kern-.025em b}\kern-.08em
    T\kern-.1667em\lower.7ex\hbox{E}\kern-.125emX}}
\begin{document}

\title{
Edge-Based Predictive Data Reduction for Smart Agriculture: A Lightweight Approach to Efficient IoT Communication
\\
\thanks{This work has been supported by the Horizon Europe WIDERA program under the grant agreement No. 101079214.}
}

\author{

    \IEEEauthorblockN{
        Dora Kreković\textsuperscript{1, *}, Mario Kušek\textsuperscript{1}, Ivana Podnar Žarko\textsuperscript{1}, Danh Le-Phuoc\textsuperscript{2}
    }
    \IEEEauthorblockA{
        \textsuperscript{1}University of Zagreb, Faculty of Electrical Engineering and Computing, Zagreb, Croatia \\
        Email: \{dora.krekovic, mario.kusek, ivana.podnar\}@fer.hr
    }
    \IEEEauthorblockA{
        \textsuperscript{2}Technical University of Berlin, Berlin, Germany \\
        Email: danh.le-phuoc@tu-berlin.de
    }
     \thanks{\textsuperscript{*}Corresponding author: Dora Kreković (dora.krekovic@fer.hr)}
}

\maketitle

\begin{abstract}
The rapid growth of IoT devices has led to an enormous amount of sensor data that requires transmission to cloud servers for processing, resulting in excessive network congestion, increased latency and high energy consumption. This is particularly problematic in resource-constrained and remote environments where bandwidth is limited, and battery-dependent devices further emphasize the problem. Moreover, in domains such as agriculture, consecutive sensor readings often have minimal variation, making continuous data transmission inefficient and unnecessarily resource intensive. To overcome these challenges, we propose an analytical prediction algorithm designed for edge computing environments and validated through simulation. The proposed solution utilizes a predictive filter at the network edge that forecasts the next sensor data point and triggers data transmission only when the deviation from the predicted value exceeds a predefined tolerance. A complementary cloud-based model ensures data integrity and overall system consistency. This dual-model strategy effectively reduces communication overhead and demonstrates potential for improving energy efficiency by minimizing redundant transmissions. In addition to reducing communication load, our approach leverages both in situ and satellite observations from the same locations to enhance model robustness. It also supports cross-site generalization, enabling models trained in one region to be effectively deployed elsewhere without retraining. This makes our solution highly scalable, energy-aware, and well-suited for optimizing sensor data transmission in remote and bandwidth-constrained IoT environments.

\end{abstract}

\begin{IEEEkeywords}
Data transmission, Data reduction, IoT, Data Prediction, Edge Computing, Smart Agriculture, Decentralized inference, Resource-constrained environments, Ag-IoT
\end{IEEEkeywords}

\section{Introduction}
\label{intro}

The demand for sustainable and data-driven agriculture is driving the adoption of intelligent monitoring systems based on the Internet of Things (IoT) and Artificial Intelligence (AI) \cite{SUSTAINABILITY}. These technologies promise improved yield productivity, efficient resource use, and better adaptation to climate variability. However, deploying such systems in remote rural farms introduces critical challenges related to connectivity, power consumption, and data availability \cite{review_agriculture}. Agro-IoT deployments typically consist of distributed low-power sensors that measure environmental parameters such as air and soil temperature, humidity, and precipitation. These devices operate on limited battery reserves and rely on wireless communication protocols, ranging from short-range (Bluetooth, Zigbee, WiFi \cite{ble, wifi, zigbee}) to long-range (LoRa, Sigfox, NB-IoT \cite{sigfox, nb-iot_lora, lora}), to transmit data to the cloud. 
While cloud-based architectures enable centralized data storage and analytics, they also create bottlenecks:
\begin{enumerate}
\item Dependence on stable Internet connectivity, often lacking in rural areas.
\item Continuous transmission of redundant or slowly changing sensor readings, leading to energy drain and network congestion.
\end{enumerate}

To address these constraints, edge computing has emerged as a promising paradigm \cite{edge_intro}. It offers a compelling alternative to centralized cloud-based solutions by bringing intelligence closer to the data source. In agricultural contexts where sensor readings change gradually, transmitting every measurement to the cloud is often unnecessary and inefficient \cite{26b9ffa541344f19976d1b133f689c0e, energy_iot}. Processing data locally on a sensor or gateway node reduces bandwidth usage, lowers latency, and preserves battery life \cite{ieee_edge23}. In the proposed use case, the edge device functions as a local gateway that collects readings from nearby environmental sensors and performs lightweight predictive inference, transmitting only significant deviations to the cloud.
Yet, a common limitation in deploying predictive models to the edge is the lack of historical labeled data required for training. This is particularly problematic in newly instrumented or under-monitored regions. To address this, we propose an edge-based prediction architecture that leverages pretrained models on publicly available satellite-derived climate datasets and transfers them to \textit{in-situ} stations without retraining. This cross-site inference strategy enables rapid deployment while maintaining high prediction accuracy, eliminating the need to train and update separate models for each location. Our approach employs a LSTM model suitable for the edge deployment. Each device forecasts future measurements and transmits data only when the observed value deviates from the prediction by more than a configurable error threshold. This predictive filtering significantly reduces the volume of transmitted data, conserving both energy and bandwidth without sacrificing data integrity.
The primary contributions of this paper are summarized as follows:
\begin{itemize}
\item \textit{Edge-Based Predictive Filtering:} We implement a compact LSTM model optimized for edge deployment, enabling local forecasting and reducing communication frequency.
\item \textit{Cross-Site Generalization:} We demonstrate that models trained on satellite-derived temperature data from one region generalize well to other in-situ locations, supporting scalable deployment in data-scarce environments.
\item \textit{Real-World Evaluation:} Using agrometeorological datasets from Croatia, we demonstrate substantial reductions in communication load (up to 94\%) while maintaining high prediction accuracy under realistic deployment conditions.
\end{itemize}
The remainder of this paper is organized as follows: Section \ref{rw} reviews related work. Section \ref{algorithm} details the prediction and transmission algorithm as well as model architecture. Section \ref{use_case} describes the datasets and deployment strategy. Section \ref{results} presents experimental results. Finally, Section \ref{conclusion} concludes the paper and outlines directions for future research.
\section{Related Work}
\label{rw}
Reducing data transmission in IoT networks is a well-studied problem, especially in domains like smart agriculture where energy efficiency and intermittent connectivity are critical constraints~\cite{constraints}. Data reduction strategies are typically categorized as compression, aggregation, or prediction-based~\cite{podjela,KREKOVIC2025101553}, each with its own trade-offs in complexity, energy usage, and fidelity.

\subsection{Data Compression and Aggregation}
Data compression reduces payload size by encoding readings at the source and decoding at the destination~\cite{NASSRA2023100806}. Techniques such as DASA~\cite{vyas2023dasa} and Huffman-based encoding~\cite{hussein2022compression} show effectiveness in edge environments. In Agricultural IoT, combining compression with secure protocols like MQTT has been explored to ensure scalability and data integrity~\cite{efficient_agri_mqtt}. However, these techniques often underperform in slowly changing environments, typical in agriculture. Aggregation methods reduce data by summarizing multiple values (e.g., averages or statistics) before transmission~\cite{aggregation_survey, multi_tier}. These methods are often implemented at gateways or cluster heads to alleviate network congestion. While scalable, they can obscure temporal trends important for fine-grained decision-making and result in loss of detail.

\subsection{Prediction-Based Data Reduction}

Prediction-based methods are an effective class of techniques for reducing data transmission in IoT systems. These methods forecast future sensor values and suppress transmission when the predicted value falls within an acceptable error range, thus minimizing communication overhead. Traditional approaches rely on statistical models such as autoregression or exponential smoothing. More recent work leverages machine learning and deep learning models, MLPs, LSTMs, CNNs, and attention-based architectures, for improved accuracy in capturing complex temporal dependencies~\cite{DURAI, putra2023efficient, morales}. Adaptive frameworks further enhance flexibility. For instance,\cite{adaptive_dual_s0925231221016295} proposed an adaptive dual prediction model that adjusts in real time to environmental dynamics.
Prediction can occur centrally~\cite{dias2016survey} or locally at the sensor node~\cite{idrees2020twolayer, hussein2023distributed}. Local prediction reduces dependency on constant connectivity but requires computationally efficient models. Dual prediction schemes (DPS)\cite{wu2020arucep, liazid2023aggregation, fathalla2022lstm, jain2022combinational} synchronize models at both the sensor and gateway, transmitting only when both predictions fail. While this improves robustness, it adds synchronization and update complexity. Extensions of DPS, such as online model updating and attention-based deep learning for long-term forecasting\cite{wu2024lidps, morales}, improve adaptability in dynamic environments, though often at higher computational cost.
In contrast, our approach distinguishes itself by using pre-trained deep learning models that can be deployed on edge devices, eliminating the need for site-specific historical data. Satellite-derived temperature data is used to train generalizable models that work across unseen locations, enabling scalable deployment and reducing transmission by over 90\% while preserving the ability to reconstruct the data stream when necessary.

\subsection{Hybrid and Domain-Specific Approaches}
In smart agriculture and environmental monitoring, hybrid strategies combine techniques like prediction, aggregation, and compression to adapt to data variability and energy constraints~\cite{ibrahim2021hybrid}. Spatial and temporal correlations have also been used to suppress redundancy: Koteich et al.\cite{spatio_temporal} exploited neighborhood relationships, while Salim et al. \cite{salim} applied Pearson correlation to reduce transmissions. These approaches work well in homogeneous environments but lose effectiveness in heterogeneous fields with varying soil and microclimates~\cite{idrees2020twolayer, mcdonnell2024agriculture}.
To address cloud limitations, fog-based systems distribute computation across nodes, edge, and cloud~\cite{vyas2024fog, anand2024fog}. LoRa-enabled platforms with local ML show promise for low-connectivity areas but often lack online updates or depend on cloud processing. As noted by Chien et al.~\cite{smart_farming_edge_s277266222200011x}, edge computing offers gains in latency and scalability but requires lightweight and adaptive models. 

\subsection{Limitations of Prior Work and Motivation}
Despite the depth of literature on data reduction, several limitations remain. Traditional compression and aggregation methods often overlook data semantics and temporal dependencies. Centralized prediction approaches rely on constant connectivity and introduce latency, while decentralized or dual prediction strategies, though more resilient, can be computationally intensive and require frequent synchronization. Moreover, many existing methods fail to generalize across heterogeneous environments typical of open-field agriculture, where sensor behavior varies due to microclimates and terrain diversity. Our approach tackles these limitations by introducing a lightweight, decentralized architecture that deploys pre-trained neural models on edge devices. A prediction-triggered transmission strategy ensures that only significant deviations are communicated, reducing network load while preserving accuracy. This design enables local adaptability, low latency, and energy-efficient operation, which are key requirements for practical Agro-IoT deployments. In terms of scalability and robustness, the proposed method is scalable because a single model, once trained using rich historical and satellite data, can be deployed across multiple sites, including new locations where no prior measurements exist. This allows for streamlined and synchronous deployment of both the model and the sensor hardware.  Robustness comes from on-device self-monitoring: by comparing predicted and observed values in real time, the system detects when predictions fail and resumes full data transmission. This feedback loop justifies the presence of field sensors, as they serve as validation checkpoints and enable reactive analytics when environmental conditions shift. To better contextualize the contributions of existing data reduction methods, Table~\ref{tab:rw_overview} provides an overview of selected state-of-the-art methods in the domain of IoT-based agriculture, outlining their domains, core strategies, evaluation datasets, and key strengths and limitations. The comparison highlights trade-offs in computational cost, adaptability, and deployment feasibility for energy-constrained agricultural IoT settings.

\begin{table*}[ht]
\centering
\renewcommand{\arraystretch}{1.2}
\small
\begin{tabular}{|p{1.3cm}|p{1.4cm}|p{1.8cm}|p{1.8cm}|p{1.6cm}|p{2cm}|p{2cm}|p{2.2cm}|}
\hline
\textbf{Article} & \textbf{Domain} & \textbf{Reduction Approach} & \textbf{Main Context} & \textbf{Dataset} & \textbf{Advantages} & \textbf{Limitations} & \textbf{Core Idea} \\ \hline


Fathalla et al. (2022) \cite{fathalla2022lstm} & IoT / WSN & DL-based Edge Prediction & Distributed edge–fog adaptive transmission & Intel Lab \& UCI Air Quality & Adaptive, accurate, real data, scalable & Static threshold, retraining cost, univariate & STM with fog-side retraining \\ \hline

Chien et al. (2022) \cite{smart_farming_edge_s277266222200011x} & Smart Farming & Edge IoT Architecture Review & Review of edge computing in agriculture & Review study & Comprehensive overview of trends & No empirical validation & Survey of edge paradigms and Ag-IoT challenges \\ \hline

Vyas et al. (2024) \cite{vyas2024fog} & Smart Agriculture & Regression-based Dual Prediction & Fog-enabled LoRa data forwarding & Smart Fasal testbed & Lightweight, low power, field-tested & Fixed threshold, limited adaptivity & Local regression sends weight updates to fog \\ \hline

Anand et al. (2024) \cite{anand2024fog} & Smart Agriculture & Analytical Estimation Prediction & Fog-assisted LoRa data reduction & Soria \& Cartagena datasets & Ultra-low power, self-adaptive & Handles slow dynamics only & Iterative estimation updates reduce LoRa traffic \\ \hline

Our Approach & Smart Agriculture & Edge DL + Adaptive Transmission & Scalable, adaptive, communication reduction  & Satellite + in situ data & High reduction, cross-site generalization & Requires model validation, sensor drift detection & Deploy pretrained DL models; transmit only on error $>$ threshold \\ \hline
\end{tabular}
\caption{Comparative overview of state-of-the-art data reduction methods in smart agriculture and IoT.}
\label{tab:rw_overview}
\end{table*}

\section{Prediction Algorithm}
\label{algorithm}
To minimize data transmission while maintaining prediction accuracy, we propose an edge-based selective communication algorithm, outlined in Algorithm~\ref{alg:prediction_transmission}. The approach relies on a pretrained time-series prediction model deployed on an edge device, which performs real-time forecasting of sensor measurements using a rolling buffer of past values.

\begin{algorithm}[h]
\caption{Edge-based Time Series Prediction and Transmission Algorithm}
\label{alg:prediction_transmission}
\begin{algorithmic}[1]
\REQUIRE Series of incoming sensor measurements $x_1, x_2, \dots, x_n$
\REQUIRE Trained prediction model $\mathcal{M}$, error threshold $\epsilon$, buffer size $k$
\STATE \textbf{Initialize:} buffer $B \gets [\,]$, transmit flag \texttt{flag} $\gets$ \texttt{True}
\FOR{each new measurement $x_t$}
    \IF{\texttt{flag} is \texttt{True}}
        \STATE \textbf{Transmit:} $x_t$
        \STATE $B \gets [x_t]$
        \STATE \texttt{flag} $\gets$ \texttt{False}
    \ELSE
        \STATE \textbf{Predict:} $\hat{x}_t \gets \mathcal{M}(B)$
        \IF{$|x_t - \hat{x}_t| > \epsilon$}
            \STATE \textbf{Transmit:} $x_t$
            \STATE $B \gets [x_t]$
        \ELSE
            \STATE $B \gets B \cup [x_t]$
            \STATE \textbf{Maintain buffer window:} $B \gets B[-k:]$
        \ENDIF
    \ENDIF
\ENDFOR
\end{algorithmic}
\end{algorithm}

At each time step $t$, the model generates a prediction $\hat{x}_t$ based on recent measurements. The actual sensor reading $x_t$ is then compared to this prediction. If the absolute error $|x_t - \hat{x}_t|$ exceeds a predefined tolerance threshold $\epsilon$, the measurement is transmitted to the cloud. Otherwise, the data is withheld, since it adds no significant information beyond the model's forecast. This strategy ensures that only informative or anomalous values are sent, preserving bandwidth and reducing energy consumption. Unlike traditional cloud-centric models that require constant connectivity and high bandwidth, this approach enables decentralized intelligence directly at the sensor level. It is particularly well-suited for constrained IoT deployments in smart agriculture, where devices operate in bandwidth-limited and power-sensitive environments. By transmitting only poorly predicted values, the system achieves a balance between communication efficiency and data fidelity, while supporting real-time monitoring and decision-making. The algorithm maintains a buffer $B$ of recent inputs to update the prediction context and adapt to temporal patterns. The overall result is a scalable and lightweight data reduction method applicable to a wide range of precision agriculture use cases.

\subsection{Model Design}

The predictive model used in this work is based on a Long Short-Term Memory (LSTM) architecture, selected for its ability to capture long-term dependencies in sequential data. LSTM networks are well-suited for time series forecasting tasks in environmental monitoring due to their capacity to model temporal patterns, seasonal effects, and autocorrelated noise. Simpler statistical approaches such as SARIMA were also tested but showed lower predictive accuracy and limited ability to capture nonlinear dependencies \cite{sst}. Our model is structured as a univariate time series forecaster. It receives as input a sequence of the 24 most recent hourly measurements for a single feature (e.g., air temperature), and it produces a one-step-ahead forecast, predicting the value for the next hour. This formulation allows the model to operate efficiently with minimal historical context while remaining sensitive to recent changes in the input signal.
The architecture comprises:
\begin{itemize}
\item An input layer receiving a sequence of shape $(24, 1)$;
\item A single LSTM layer with 64 hidden units, followed by a dropout layer (rate = 0.2) for regularization;
\item A fully connected (dense) output layer with linear activation, producing a scalar prediction.
\end{itemize}

Training was conducted using the mean squared error (MSE) loss function and the Adam optimizer with an initial learning rate of 0.001. Early stopping and learning rate decay were applied to prevent overfitting and accelerate convergence. The model was trained for up to 100 epochs with a batch size of 32. With approximately 10,000 trainable parameters and a single recurrent layer, the model remains computationally efficient, requiring minimal memory and enabling real-time inference on resource-constrained edge devices. 
The implementation is compatible with both TensorFlow Lite and ONNX Runtime, allowing flexible deployment across various hardware platforms. This design enables localized prediction directly at the sensor node, thereby reducing the need for continuous data transmission to centralized servers. When combined with our edge-triggered transmission logic, this contributes to substantial energy and bandwidth savings in field deployments.

\section{Use Case and Model Deployment}
\label{use_case}

Many agricultural regions lack sufficient in-situ sensor coverage, which poses challenges for training data-driven forecasting models. To address this limitation, our approach leverages freely available satellite data as a proxy for ground measurements, enabling model development and deployment even in data-scarce environments. The framework targets smart agriculture applications such as frost protection, where timely responses to environmental changes are crucial for crop health and resource efficiency \cite{agronomy15051164}.

\subsection{Datasets}
\label{subsec:datasets}

We use two complementary datasets for model training and evaluation: the Copernicus ERA5-Land dataset and the PinovaMeteo in-situ sensor network~\cite{rs16040641}. The ERA5-Land dataset (2019–2021) provides hourly gridded environmental variables at high spatial and temporal resolution across Europe \cite{era5land}, offering broad coverage and consistent historical context, which makes it ideal for model training, especially in regions with limited in-situ measurements. In parallel, the PinovaMeteo network, a distributed system of agrometeorological stations across Croatia, provides 10-minute resolution measurements.
The study ~\cite{rs16040641} shows strong correlations, particularly for temperature, between satellite-based ERA5-Land data and ground-based measurements in Croatia, supporting the use of satellite data for proxy training. By integrating both datasets, our approach combines large-scale environmental context with fine-grained local accuracy, enabling robust inference in both in-situ and cross-site settings. Temperature serves as the primary variable for the model; however, the framework can easily be extended to other sensor parameters such as humidity, soil moisture, or rainfall.

\subsection{Deployment Scenarios and Model Transferability}
\begin{figure}[t!]
    \centering
    \includegraphics[width=\columnwidth]{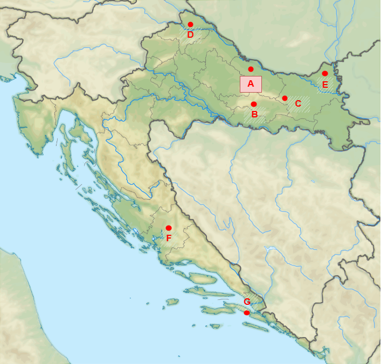}
    \caption{Sensor locations across Croatia.}
    \label{map_sites}
\end{figure}

To explore the robustness and transferability of our approach, we designed a series of model training and inference scenarios across a network of geographically distributed stations:
\begin{itemize}
    \item \textit{Same-site training and testing:} Models are trained and tested on \textit{in-situ} data from the same location. This represents an idealized setting that may be infeasible in low-resource environments.
    \item \textit{Cross-site inference:} Models trained at one location using \textit{in-situ} readings are applied to other sites without retraining. This reflects realistic deployment conditions in which labeled training data may only be available at a subset of nodes. We assess whether such transferred models can still provide useful predictions and how spatial proximity or environmental similarity influences performance.    
    \item \textit{Satellite-trained, same-site inference:} Models are trained solely on satellite data from a given location and evaluated against \textit{in-situ} measurements at the same site. This simulates conditions where ground truth is unavailable during training, reflecting data-scarce environments.
    \item \textit{Satellite-trained, cross-site inference:} Models are trained on satellite data from one location and used to predict \textit{in-situ} values at a different location. This tests both data-type and spatial transferability, relevant for scaling models in data-scarce regions.
\end{itemize}
Through these experiments, we aim to understand under which conditions model transfer is beneficial across sites and using different sources of data. Figure~\ref{map_sites} shows the spatial distribution of selected sensor locations across Croatia, labeled A–G. Table~\ref{tab:location_distances} summarizes inter-site distances, which serve as a reference for interpreting transfer performance relative to geographic separation. We analyze model behavior across different source–target pairs using two prediction error thresholds (0.5 and 1.0°C), which are commonly used in agricultural monitoring as decision points for triggering sensor measurements \cite{thresholds}.
\begin{table}[htbp]
\centering
\caption{Distances between sensor locations (in km)}
\label{tab:location_distances}
\begin{tabular}{|c|c|c|c|c|c|c|}
\hline
\textbf{From / To} & \textbf{B} & \textbf{C} & \textbf{D} & \textbf{E} & \textbf{F} & \textbf{G} \\ \hline
\textbf{A} & 50 & 58 & 79.5 & 94 & 233 & 318 \\ \hline
\end{tabular}
\end{table}

\section{Results and Evaluation}
\label{results}
This section presents a detailed evaluation of our proposed approach using two complementary datasets introduced in Section~\ref{subsec:datasets}: the Copernicus ERA5-Land dataset and the PinovaMeteo dataset from in-situ stations. While ERA5-Land offers broad spatial coverage and historical context through hourly gridded data, PinovaMeteo provides high-resolution, real-time measurements from local stations. This combination allows us to assess both generalization across regions and fine-grained performance in localized settings. 
We assess model performance using two primary metrics: (i) \textbf{Data Reduction} and (ii) \textbf{Mean Absolute Error (MAE)}. In our threshold-triggered system, prediction accuracy and data reduction are numerically equivalent by design, since correctly predicted values (within the defined threshold) are not transmitted. Thus, data reduction serves as a direct indicator of communication efficiency. On the other hand, MAE quantifies the average absolute difference between the predicted and actual sensor values, regardless of whether the data point was transmitted or not. This metric offers a threshold-independent view of model accuracy and is essential for assessing the underlying quality of the predictions. By reporting MAE, we ensure that even small but systematic deviations are captured.

\begin{enumerate}
    \item \textbf{Data Reduction:} Quantifies the percentage of sensor readings which are not transmitted due to successful prediction within the specified threshold:
    \begin{equation}
        \text{Data Reduction (\%)} = \left( 1 - \frac{\text{Data Transmitted}}{\text{Total Data}} \right) \times 100
        \label{eq:data_reduction}
    \end{equation}
    \begin{itemize}
        \item \textit{Total Data:} Total number of data points generated by the sensor station.
        \item \textit{Data Transmitted:} Number of samples actually transmitted, based on prediction-triggered filtering.
    \end{itemize}

    \item \textbf{Mean Absolute Error (MAE):} Represents the average magnitude of prediction errors:
    \begin{equation}
        \text{MAE} = \frac{1}{N} \sum_{i=1}^{N} \left| \hat{y}_i - y_i \right|
    \end{equation}
    where $\hat{y}_i$ is the predicted value, $y_i$ is the ground truth, and $N$ is the total number of samples.
\end{enumerate}

\begin{table*}[t!]
\centering
\caption{\textit{In-situ} training and inference: Data transmission reduction at each site using models trained on \textit{in-situ} data.}
\label{tab:scenario1}
\renewcommand{\arraystretch}{1.4}
\begin{tabular}{|c|c||r|r|r|c||r|r|r|c|}
\hline
\textbf{Location} & \textbf{MAE} 
& \multicolumn{4}{c||}{\textbf{Threshold = 0.5°C}} 
& \multicolumn{4}{c|}{\textbf{Threshold = 1.0°C}} \\
\cline{3-10}
& 
& \textbf{Total} & \textbf{Correct} & \textbf{Transmitted  Samples} & \textbf{Reduction (\%)} 
& \textbf{Total} & \textbf{Correct} & \textbf{Transmitted  Samples} & \textbf{Reduction (\%)} \\
\hline
A & 0.264 & 52535 & 45725 & 6810  & 87.04 & 52535 & 51260 & 1275  & 97.57 \\ \hline
B & 0.269 & 52535 & 45962 & 6573  & 87.49 & 52535 & 51346 & 1189  & 97.74 \\ \hline
C & 0.287 & 52535 & 44632 & 7903  & 84.96 & 52535 & 50809 & 1726  & 96.71 \\ \hline
D & 0.255 & 52535 & 46122 & 6413  & 87.79 & 52535 & 51533 & 1002  & 98.09 \\ \hline
E & 0.316 & 52535 & 42906 & 9629  & 81.67 & 52535 & 50407 & 2128  & 95.95 \\ \hline
F & 0.347 & 52535 & 40661 & 11874 & 77.40 & 52535 & 49409 & 3126  & 94.05 \\ \hline
G & 0.324 & 52535 & 43519 & 9016  & 82.84 & 52535 & 50418 & 2117  & 95.97 \\ \hline
\end{tabular}
\end{table*}

\begin{table*}[t!]
\centering
\renewcommand{\arraystretch}{1.4}
\caption{Cross-site in-situ inference: Generalization performance of a model trained at Location A on \textit{in-situ} data (Source) and evaluated at other locations using obtained \textit{in-situ} measurements (Target).}
\label{tab:scenario2}
\begin{tabular}{|c|c|c||r|r|r|c||r|r|r|c|}
\hline
\textbf{Source (\textit{In-situ})} & \textbf{Target (\textit{In-situ})} & \textbf{MAE} 
& \multicolumn{4}{c||}{\textbf{Threshold = 0.5°C}} 
& \multicolumn{4}{c|}{\textbf{Threshold = 1.0°C}} \\
\cline{4-11}
& & 
& \textbf{Total} & \textbf{Correct} & \textbf{Transmitted} & \textbf{Reduction (\%)} 
& \textbf{Total} & \textbf{Correct} & \textbf{Transmitted} & \textbf{Reduction (\%)} \\
\hline
A & B & 0.291 & 52535 & 45654 & 6881 & 86.90 & 52535 & 51230 & 1305 & 97.52 \\ \hline
A & C & 0.292 & 52535 & 43933 & 8602 & 83.63 & 52535 & 50587 & 1948 & 96.29 \\ \hline
A & D & 0.260 & 52535 & 45943 & 6592 & 87.45 & 52535 & 51481 & 1054 & 97.99 \\ \hline
A & E & 0.297 & 52535 & 43967 & 8568 & 83.69 & 52535 & 50717 & 1818 & 96.54 \\ \hline
A & F & 0.308 & 52535 & 42876 & 9659 & 81.61 & 52535 & 50485 & 2050 & 96.10 \\ \hline
A & G & 0.309 & 52535 & 44360 & 8175 & 84.44 & 52535 & 50903 & 1632 & 96.89 \\ \hline
\end{tabular}
\end{table*}

\subsubsection{In-situ Training and Inference (Same Location)}
In the baseline scenario, models were trained and evaluated using \textit{in-situ} data from the same agrometeorological station. Table~\ref{tab:scenario1} demonstrates that the models consistently achieved high predictive performance across all locations. At a threshold of 0.5\,°C, the models achieved MAE between 0.255 and 0.347. The corresponding data reduction ranged from 77.40\% to 87.79\%, indicating that the majority of sensor readings could be accurately predicted and therefore skipped from transmission. When the threshold was increased to 1.0\,°C, data reduction improved significantly, ranging from 94.05\% to 98.09\%. This demonstrates that relaxing the error tolerance allows the system to suppress even more transmissions without significant degradation of prediction quality. These results indicate that when inference is performed at the same site where the training data was collected, the models are highly effective at both accurate prediction and data reduction.

\subsubsection{Cross-Site In-situ Inference}
This scenario evaluates the generalization capability of a model trained at a single site (Location A) when applied to unseen \textit{in-situ} data from six other locations (B to G). As shown in Table~\ref{tab:scenario2}, the model achieved strong predictive performance across all target sites. At a threshold of 0.5\,°C, the data reduction ranged from 81.61\% to 87.45\%, with MAE between 0.260 and 0.309. Increasing the threshold to 1.0\,°C further improved communication efficiency, reducing transmitted samples by over 96\% at all locations while maintaining low MAE values. These results suggest that a single \textit{in-situ}-trained model can be effectively generalized to multiple target sites, including those with both similar and varying agroclimatic conditions (see Figure~\ref{map_sites}), thereby reducing the need for location-specific model retraining and simplifying large-scale deployment.

\begin{table*}[t!]
\centering
\caption{Satellite-to-in-situ inference: Generalization performance of a model trained at Location A on satellite data and evaluated at same locations using obtained \textit{in-situ} measurements (Location).}
\label{tab:scenario3}
\renewcommand{\arraystretch}{1.3}
\begin{tabular}{|c|c||r|r|r|c||r|r|r|c|}
\hline
\textbf{Location} & \textbf{MAE} 
& \multicolumn{4}{c||}{\textbf{Threshold = 0.5°C}} 
& \multicolumn{4}{c|}{\textbf{Threshold = 1.0°C}} \\
\cline{3-10}
& 
& \textbf{Total} & \textbf{Correct} & \textbf{Transmitted Samples} & \textbf{Reduction (\%)} 
& \textbf{Total} & \textbf{Correct} & \textbf{Transmitted Samples} & \textbf{Reduction (\%)} \\
\hline
A & 0.395 & 8735 & 6474 & 2261 & 74.12 & 8735 & 8059 & 676 & 92.26 \\ \hline
B & 0.399 & 8735 & 6386 & 2349 & 73.11 & 8735 & 8037 & 698  & 92.01 \\ \hline
C & 0.386 & 8735 & 6527 & 2208 & 74.73 & 8735 & 8075 & 660  & 92.44 \\ \hline
D & 0.376 & 8735 & 6541 & 2194 & 74.87 & 8735 & 8148 & 587  & 93.29 \\ \hline
E & 0.373 & 8735 & 6595 & 2140 & 75.50 & 8735 & 8158 & 577  & 93.41 \\ \hline
F & 0.377 & 8735 & 6671 & 2064 & 76.36 & 8735 & 8085 & 650  & 92.57 \\ \hline
G & 0.360 & 8735 & 6636 & 2099 & 75.95 & 8735 & 8228 & 507  & 94.21 \\ \hline
\end{tabular}
\end{table*}

\begin{table*}[t!]
\centering
\caption{Satellite-to-in-situ inference (cross-site): Generalization performance of a model trained at Location A on satellite data (Source) and evaluated at other locations using obtained \textit{in-situ} measurements (Target).}
\label{tab:scenario4}
\renewcommand{\arraystretch}{1.3}
\begin{tabular}{|c|c|c||r|r|r|c||r|r|r|c|}
\hline
\textbf{Source (Satellite)} & \textbf{Target (In-situ)} & \textbf{MAE}
& \multicolumn{4}{c||}{\textbf{Threshold = 0.5}} 
& \multicolumn{4}{c|}{\textbf{Threshold = 1.0}} \\
\cline{4-11}
& & 
& \textbf{Total} & \textbf{Correct} & \textbf{Transmitted} & \textbf{Reduction (\%)} 
& \textbf{Total} & \textbf{Correct} & \textbf{Transmitted} & \textbf{Reduction (\%)} \\
\hline
A & B & 0.484 & 8735 & 5720 & 3015 & 65.48 & 8735 & 7731 & 1004 & 88.51 \\ \hline
A & C & 0.390 & 8735 & 6488 & 2247 & 74.28 & 8735 & 8092 & 643 & 92.64 \\ \hline
A & D & 0.405 & 8735 & 6347 & 2388 & 72.66 & 8735 & 8063 & 672 & 92.31 \\ \hline
A & E & 0.525 & 8735 & 5069 & 3666 & 58.03 & 8735 & 7548 & 1187 & 86.41 \\ \hline
A & F & 0.390 & 8735 & 6599 & 2136 & 75.55 & 8735 & 8007 & 728 & 91.67 \\ \hline
A & G & 0.459 & 8735 & 5713 & 3022 & 65.40 & 8735 & 7877 & 858 & 90.18 \\ \hline
\end{tabular}
\end{table*}

\subsubsection{Satellite-to-In-situ Inference (Same Site)}
In this setting, models trained on satellite-derived data were tested against the ground-truth \textit{in-situ} observations at the same geographic locations. This simulates an early deployment phase where no ground-based data is yet available. Table~\ref{tab:scenario3} shows that, even without access to local sensor data during training, the models achieved reasonable accuracy. At the 0.5\,°C threshold, data reduction ranged from 73.11\% to 76.36\%, and MAE values remained low (0.360 to 0.399). Raising the threshold to 1.0\,°C led to over 92\% reduction in transmissions at all locations, confirming that satellite-trained models can be practical for initial inference with moderate error tolerance, particularly when moderate prediction error is acceptable.

\subsubsection{Satellite-to-In-situ Cross-Site Inference}
The final scenario examined whether a satellite-trained model from Location A could generalize to entirely different \textit{in-situ} locations (B to G). Table~\ref{tab:scenario4} highlights broader variability in performance, reflecting greater domain shift. At the 0.5\,°C threshold, data reduction ranged from 58.03\% to 75.55\%, with MAE values between 0.390 and 0.525. Performance improved significantly at the 1.0\,°C threshold, achieving over 86\% reduction in all cases and up to 92.64\%. While results were less consistent than in the same-site scenario, the model still delivered useful predictions, suggesting that satellite pretraining can enable rapid, data-free deployment in new locations, especially when used with relaxed error thresholds.

\subsection{Summary}

Our results demonstrate that:
\begin{itemize}
    \item Models trained on high-resolution \textit{in-situ} data achieve high accuracy and data reduction when used at the same site.
    \item Cross-site inference using \textit{in-situ}-trained models maintains strong generalizability, reducing the need to retrain the model for each new location.
    \item Satellite-trained models offer a practical alternative where on-site data is unavailable, though they perform with lower accuracy compared to models trained on \textit{in-situ} measurements.
    \item Across all scenarios, raising the error threshold improves both accuracy and transmission efficiency.
\end{itemize}

These findings highlight the value of predictive filtering in edge-AI deployments, especially for precision agriculture where bandwidth and energy are limited. Satellite data can support early deployment, while \textit{in-situ} training enhances long-term performance as local data becomes available.

\section{Conclusion and Future Work}
\label{conclusion}

The massive volume of data generated by IoT devices in smart agriculture necessitates intelligent transmission strategies that go beyond traditional cloud-centric models. In this paper, we propose an edge-enabled prediction framework using LSTM models to reduce redundant data transmissions to the cloud while allowing for accurate reconstruction of the original data stream. Our approach addresses the limitations of bandwidth-constrained, intermittently connected environments typical of agricultural deployments. By processing data locally at the edge and only transmitting measurements that deviate beyond a defined error threshold, we can significantly reduce the communication load. Our experiments demonstrate that data reduction even higher than 90\% can be achieved in many cases, without sacrificing data quality.  The proposed approach enables real-time, energy-efficient monitoring even in connectivity-constrained environments.
Cross-site inference results confirmed the generalizability of trained models across geographically diverse regions in Croatia. Under moderate error thresholds, our system consistently achieved over 92\% prediction accuracy while reducing transmitted data by up to 94\%. This highlights the potential of satellite data as a viable proxy for training in regions where in-situ measurements are limited or unavailable. Compared to conventional data reduction techniques that rely on compression or central prediction, our approach is more adaptive, scalable, and suited for field deployment. It ensures local autonomy while minimizing reliance on cloud connectivity, a critical feature for rural agricultural use cases.

While the presented results are promising, several challenges remain. One important direction for future work is optimizing model size for deployment on ultra-low-power platforms, such as microcontrollers or solar-powered embedded devices. Evaluating and reducing the memory and computational footprint will be critical for widespread edge deployment. Although the current evaluation was performed in a simulated setting, the model’s compact design indicate its feasibility for deployment on resource-constrained IoT hardware. Future work will include empirical validation of energy consumption and latency through deployment on edge devices. Another avenue involves developing mechanisms for online model adaptation. By enabling periodic retraining or even on-device learning, the system could dynamically adjust to changing environmental conditions, improving long-term reliability. Additionally, reducing the cost of retraining through incremental or federated learning approaches could significantly lower bandwidth requirements while preserving data privacy. Expanding the framework to support multivariate and multi-modal inference, by incorporating additional environmental variables such as soil moisture, rainfall, or humidity, could also enhance the robustness and accuracy of the predictions. Moreover, future work could further extend the proposed approach by introducing adaptive thresholding mechanisms, allowing the system to adjust temperature-based decision limits dynamically rather than relying on fixed values. Finally, long-term deployment studies are needed to validate energy consumption patterns and system reliability under real-world operational conditions. Overall, this work sets the foundation for intelligent, energy-efficient, and scalable smart agriculture systems capable of adapting to both environmental variability and infrastructure limitations.

\bibliographystyle{IEEEtran}
\bibliography{IEEEabrv,references}

\end{document}